\pdfoutput=1

\documentclass[11pt]{article}

\usepackage[preprint]{acl}

\usepackage{times}
\usepackage{latexsym}

\usepackage[T1]{fontenc}

\usepackage[utf8]{inputenc}

\usepackage{microtype}

\usepackage{inconsolata}

\usepackage{graphicx}

\usepackage{graphicx}  
\usepackage{caption}
\usepackage{subfigure}  
\usepackage{booktabs}  
\usepackage{multicol}
\usepackage{multirow}
\usepackage{amsfonts}
\usepackage{amsmath}
\usepackage{amssymb}
\usepackage{amstext}
\usepackage{amsthm}

\usepackage{xcolor}         
\usepackage[capitalize]{cleveref}
\usepackage{amsmath}
\usepackage{breqn}
\usepackage{bbm}
\usepackage{amsthm}
\usepackage{amssymb}
\usepackage{mathtools}
\usepackage{xspace}
\usepackage{enumitem}
\usepackage[hang,flushmargin]{footmisc}
\usepackage{wrapfig}
\setlist{leftmargin=*}
\setlist[itemize]{noitemsep}
\setlist[enumerate]{noitemsep}

\usepackage{listings}
\usepackage{framed}

\newcommand{\predSeq}{\mathbf{s}}
\newcommand{\methodname}{CSL\xspace}
\newcommand{\methodnamePrompt}{\texttt{CSL}\xspace}
\newcommand{\methodnameNext}{\texttt{CSL-Next}\xspace}

\newcommand{\baselinePTrue}{\texttt{P(true)}\xspace}
\newcommand{\baselineSAR}{\texttt{TokenSAR}\xspace}
\newcommand{\baselineNLLNorm}{\texttt{SL}(norm)\xspace}
\newcommand{\baselineNLL}{\texttt{SL}\xspace}

\newcommand{\baselineSE}{\texttt{SE}\xspace}
\newcommand{\baselineSENorm}{\texttt{SE(norm)}\xspace}
\newcommand{\methodSEAttn}{\texttt{SE+CSL}\xspace}

\newcommand{\datasetcoqa}{\texttt{coqa}\xspace}
\newcommand{\datasetnqopen}{\texttt{nq}\xspace}
\newcommand{\datasettrivia}{\texttt{trivia}\xspace}

\newcommand{\baselineDegree}{\texttt{Deg}\xspace}

\newcommand{\ConfMethod}{\text{C}_{CSL}\xspace}
\newcommand{\Conf}{\text{C}\xspace}
\newcommand{\Acc}{acc\xspace}

\newcommand{\LLaMATwoName}{LLaMA2-70B\xspace}
\newcommand{\GPTName}{\texttt{gpt-3.5-turbo-0125}\xspace}

\newcommand\attnplus[1]{{\color{red}#1}}

\definecolor{codegreen}{rgb}{0,0.6,0}
\definecolor{codegray}{rgb}{0.5,0.5,0.5}
\definecolor{codepurple}{rgb}{0.58,0,0.82}
\definecolor{backcolour}{rgb}{0.95,0.95,0.92}
\lstdefinestyle{mystyle}{
    backgroundcolor=\color{backcolour},   
    commentstyle=\color{codegreen},
    keywordstyle=\color{magenta},
    numberstyle=\tiny\color{codegray},
    stringstyle=\color{codepurple},
    basicstyle=\ttfamily\footnotesize,
    breakatwhitespace=false,         
    breaklines=true,                 
    captionpos=b,                    
    keepspaces=true,                 
    numbers=left,                    
    numbersep=5pt,                  
    showspaces=false,                
    showstringspaces=false,
    showtabs=false,                  
    tabsize=2,
commentstyle=\color{blue},
morecomment=[l]{$},
}
\title{Contextualized Sequence Likelihood: Enhanced Confidence Scores for Natural Language Generation}

\author{
 \textbf{Zhen Lin\textsuperscript{1}},
 \textbf{Shubhendu Trivedi},
 \textbf{Jimeng Sun\textsuperscript{1,2}}
\\
 \textsuperscript{1}University of Illinois at Urbana-Champaign\\
 \textsuperscript{2}Carle’s Illinois College of Medicine, University of Illinois at Urbana-Champaign,
\\
 \small{
   \textbf{Correspondence:} \href{mailto:email@domain}{zhenlin4@illinois.edu}
 }
}

\begin{document}
\maketitle
\begin{abstract}
The advent of large language models (LLMs) has dramatically advanced the state-of-the-art in numerous natural language generation tasks.
For LLMs to be applied reliably, it is essential to have an accurate measure of their confidence.
Currently, the most commonly used confidence score function is the likelihood of the generated sequence, which, however, conflates semantic and syntactic components.
For instance, in question-answering (QA) tasks, an awkward phrasing of the correct answer might result in a lower probability prediction. 
Additionally, different tokens should be weighted differently depending on the context.
In this work, we propose enhancing the predicted sequence probability by assigning different weights to various tokens using attention values elicited from the base LLM.
By employing a validation set, we can identify the relevant attention heads, thereby significantly improving the reliability of the vanilla sequence probability confidence measure.
We refer to this new score as the Contextualized Sequence Likelihood (\methodname).
\methodname is easy to implement, fast to compute, and offers considerable potential for further improvement with task-specific prompts.
Across several QA datasets and a diverse array of LLMs, \methodname has demonstrated significantly higher reliability than state-of-the-art baselines in predicting generation quality, as measured by the AUROC or AUARC.
The code to replicate our experiments is available at \url{https://github.com/zlin7/ContextSL}.
\end{abstract}

\section{Introduction}\label{sec:intro}
The development of large language models (LLMs) has afforded tremendous advancements in natural language generation (NLG). 
Recently, LLMs have been widely applied across various natural language domains \citep{zhang2023sentiment,wang2023gpt,alves-etal-2023-steering,zhang-etal-2024-la-ucl}, even extending to tasks and domains traditionally dominated by other machine learning algorithms, such as graph data~\citep{fatemi2024talk}, tabular data~\citep{borisov2023language, hegselmann2023tabllm}, time series~\citep{gruver2023llmtime, rasul2024lagllama}, predictive chemistry~\citep{Jablonka2024LeveragingLL, shi2023relm}, computer vision~\citep{wang2023visionllm}, amongst others.
As LLMs continue to demonstrate outstanding performance, their reliability is increasingly scrutinized. 
Uncertainty quantification, an area of research that can provide some guidance on reliability, has recently gained much attention. 
Despite its long history in other machine learning tasks \citep{gawlikowski2023survey}, our understanding of uncertainty quantification in NLG remains relatively limited. 

During pre-training, (autoregressive) LLMs are optimized to predict high logits for the target token to minimize (variants of) the negative log likelihood. 
Consequently, one of the most natural and widely used confidence scores in selective NLG, conformal NLG, or uncertainty quantification is the (sometimes normalized) likelihood of the sequence, equivalent to the sum/mean of token logits. 
At first glance, sequence likelihood appears to be the most faithful reflection of a model's confidence, as it represents the (log of) the predicted probability of the output sequence, or $\log{\hat{p}(\predSeq|x)}$. 
However, this measure often lacks proper contextualization and disregards the specific nature of the task at hand. 
$\hat{p}(\predSeq|x)$ conflates syntactic and semantic likelihoods, though we typically prioritize the semantic aspect, to a varying degree depending on the task. 
For example, depending on whether the question is ``Which country won the World Cup in 2022'' or ``When did Messi win the World Cup,'' the answer ``Messi emerged victorious in the 2022 World Cup'' could be considered correct or incorrect. 
Further, the somewhat unusual expression here could adversely impact $\hat{p}(\predSeq|x)$.

Despite its limitations, relatively limited attention has been paid to improving the vanilla sequence likelihood. 
A recent study by ~\citet{duan2023shifting} proposed assessing token relevance using an external natural language inference (NLI) model. 
The relevance score of a token is negatively proportional to the similarity between the original sequence and the sequence with this token removed, which is then applied to weight the token logits.
However, since modern LMs rely on sub-word tokenizers, removing one sub-word token at a time results in non-words, is computationally expensive for longer texts, and remains context-unaware.


In this paper, we investigate the potential of utilizing the LLM's own attention mechanism to develop a weighted sequence likelihood as an enhanced confidence score. 
Specifically, the LLM is prompted to concentrate on the relevant tokens in its own generation. 
The most appropriate attention heads are then employed to re-weight the original token logits. The main contributions of this paper are summarized as thus:
\begin{itemize}

\item We introduce a straightforward yet effective method to reweight token logits, resulting in \underline{C}ontextualized \underline{S}equence \underline{L}ikelihood (\methodname), a more reliable confidence measure.
        
\item We improve current automatic evaluation methods for confidence measures on Question-Answering (QA) datasets and manually verify their effectiveness.

\item In popular free-form QA datasets, and on a variety of LLMs, we verify that \methodname significantly outperforms baselines.
Case studies suggest the attention weights are meaningful.
\end{itemize}

\section{Related Works}\label{sec:related}
With the rapid proliferation of LLMs and their swift adoption across various domains, uncertainty quantification (UQ) for natural language generation (NLG) is a fast-growing area of research (see~\citet{baan2023uncertainty} and references therein). 
Adopting the language used by \citet{lin2023generating}, uncertainty measures the predictive distribution, while confidence (our focus) further depends on the specific generation.
A common approach involves reducing NLG to a de facto classification problem and utilizing or enhancing classical UQ methods for classification \citep{desai-durrett-2020-calibration,jiang-etal-2021-know,kamath-etal-2020-selective,wang-etal-2022-uncertainty,xiong2023llms}.
Recognizing the unique challenges posed by the inherently high (potentially infinite) dimensionality of NLG, recent research increasingly considers UQ for NLG from a sequence perspective~\citep{hou2023decomposing,kuhn2023semantic,malinin2021uncertainty,lin2023generating}.

Relatively less attention has been devoted to confidence measures. Sequence likelihood, or the log-probability of the generated sequence, remains one of the most popular proxies for assessing the quality of individual answers \citep{quach2023conformal,kuhn2023semantic,cole2023selectively}. Another natural approach, given the versatility and strong performance of LLMs, involves using prompts to elicit the LLM's own confidence level \cite{kadavath2022language,Lin2022TeachingMT,DBLP:journals/corr/abs-2012-14983,chen2023quantifying,he2023investigating,li2024think,wightman2023strength}.
For free-form generation datasets, this is typically done by arranging answers as options and extracting the model's logits for each option. 
While our method also employs prompts, they primarily guide the model to focus on tokens relevant to the context of the question. 
Another approach involves sampling additional generations and comparing their similarities~\citep{lin2023generating,cole2023selectively}, which demonstrates good discriminative capability but can be rather expensive. Although this approach has the advantage that it can work for black-box LLMs.
Ensemble methods have also been proposed~\citep{chen2023quantifying}.

Recently, \citet{duan2023shifting} proposed an improvement to sequence likelihood by weighting the tokens using their importance, which is computed by removing one token at a time from the original sequence and computing the NLI dissimilarity between the new and original sequences.
The removal of sub-word tokens, however, introduces drastically grammatically incorrect sentences, which could confuse the NLI model.
In addition, such a method could incur high computation overhead by making $n$ NLI comparisons, where $n$ is the length of the generation. 
In contrast, our method incurs no artificial grammatical errors and miniscule overhead. 

An important downstream application of confidence measures is \textbf{abstention in LLMs}.
Classification with rejection~\cite{Corbiere2019AddressingConfidence,Fumera2000RejectThresholds,Geifman2017SelectiveNetworks,Jiang2018ToClassifier,lin2022scrib} could be considered the direct antecedent of selective NLG---both aiming to determine when to trust a model. 
Naturally, approaches similar to those in the classification with rejection literature have been applied to NLP applications~\cite{varshney-etal-2022-investigating,varshney-etal-2022-towards}.
More recently, \citet{cole2023selectively,yadkori2024mitigating,lin2023generating,quach2023conformal} started investigating NLG with abstention from UQ or risk control perspectives.


Calibration is another crucial and relevant topic that has been extensively studied \cite{mielke-etal-2022-reducing,si-etal-2022-examining,xiong2023llms,zhu2023on}.
Although calibration is not directly related to our primary focus---our main interest lies in ranking the confidence of different measures---we include results on the calibrated performance in the Appendix.
Conformal prediction~\cite{vovk2005algorithmic} could also be considered a form of calibration at the distribution level, and has been extended to NLG to bound variants of error rates~\cite{quach2023conformal,yadkori2024mitigating}.




\section{Contextualized Sequence Likelihood}\label{sec:method}
\subsection{Background: Sequence Likelihood}

In this section we describe our approach: contextualized sequence likelihood (\methodname). 
We first fix notation and introduce relevant definitions. 
We will focus on auto-regressive LMs, the current dominant paradigm.
Denoting the model as $\mathcal{M}$, for any given input prompt denoted as $x$, the response $\predSeq$ is a sequence of tokens $[s_1,\ldots,s_n]$ sampled from the predictive distribution $P(S;x,\mathcal{M})$.
We will denote $\predSeq_{<i}$ as the truncated sequence $[s_1,\ldots,s_{i-1}]$.
Given the auto-regressive assumption, the (log-softmax'd) logit for the $i$-th token represent $\mathcal{M}$'s prediction of the log probability of token $s_i$ at this location. 
Consequently, the sum of all the logits for the sequence represents the log of the model's predicted probability of the output sequence:
\vspace{-3mm}
\begin{align}
    \Conf_{\baselineNLL} = \sum_{i=1}^n l_i = \log{\prod_{i=1}^n \hat{p}(s_i|\predSeq_{<i},x)}\label{eq:confll}.
\end{align}
\cref{eq:confll} is commonly used as a confidence score~\cite{quach2023conformal,cole2023selectively}, and often normalized by the length $n$, as otherwise longer sequences tend to receive lower confidence~\citep{kuhn2023semantic,malinin2021uncertainty}:
\vspace{-3mm}
\begin{align}
    \text{C}_{\baselineNLLNorm} = \frac{1}{n} \sum_{i=1}^n l_i\label{eq:confll:norm}.
\end{align}
As pointed out by \citet{cole2023selectively}, in practice $\text{C}_{LL}$ is far from the actual log-probability of the sequence $\mathbf{s}$, $\log{\hat{p}(\predSeq|x)}$.
Techniques like nucleus sampling or top-k sampling will reduce the sum of the predicted probability of all tokens below 1. 
However, if we view this new distribution that $\predSeq$ is effectively sampled from as $P'(S|x,\mathcal{M})$, then we at least have:
\begin{align}
    \forall \predSeq, {P'}(\predSeq|x) \propto \prod_{i=1}^n \hat{p}(s_i|\predSeq_{<i},x) \label{eq:propto}.
\end{align}
Thus, $\Conf_{\baselineNLL}$ still faithfully preserves the ranking of LM's predicted probability of all possible generated sequences, which is all we care about for a good (pre-calibrated) confidence measure.
However, as we shall see next, this does not imply sequence likelihood is a good confidence measure.


\subsection{Contextualized Likelihood via Attention}

While intuitively natural as confidence measures, \cref{eq:confll,eq:confll:norm} sweep a crucial consideration under the proverbial rug: While sequence-likelihood reflects the model's predicted probability of the sequence $\predSeq$, what does this probability actually \textit{mean}?
Unfortunately, there is an inherent ambiguity here.
For instance, let's say we are classifying an image $x$ from ImageNet~\cite{deng2009imagenet}, and take the first-choice confidence score $\hat{p}(y|x)$ as predicted by a ResNet~\cite{he2016deep}.
However, $\hat{p}(y=\text{cat}|x)$ could mean the (model's predicted) probability of ``the input image is that of a cat'', ``there is a cat in the input image'', or probably more precisely speaking, ``this image should be labeled as a cat by ImageNet standards''\footnote{See a similar discussion in~\citet{beyer2020we}.}.
Similarly, strictly speaking, $\hat{p}(\predSeq|x)$ only reflects the LM's prediction of the probability of ``$\mathbf{s}$ follows $x$, according to the training data''.

To illustrate further, consider the question-answer pair where $x$ is ``Q: What did Neil Armstrong do on July 20, 1969?''.
A response $\mathbf{s}$ which goes ``A: On July 20, 1969, Armstrong and Buzz Aldrin landed on the Moon for the first time in human history.''  appears appropriate and highly probable.
However, the confidence of ``whether $\predSeq$ correctly answers $x$'' is minimally influenced by the redundant mention of the date or Armstrong's fellow astronaut, Buzz Aldrin.
Generally speaking, the confidence that we are concerned with in a response $\predSeq$ depends on the context, with some tokens being significantly more relevant than others.

How can we systematically identify the most relevant tokens?
To address this, we propose using a prompt (\cref{fig:prompt_short}) to elicit the LM's attention on its own response $\predSeq$.
Essentially, we prompt the LM to assess whether its generated response correctly answers the question. 
Unlike previous approaches that rely on similar prompts, we disregard the actual judgment and instead extract the attention values of the LM on $\predSeq$ during this process to reweight the token logits.
Assuming the $\predSeq_i$ becomes the $i'$-th token in the attention-eliciting prompt, the new confidence score could be written as:
\begin{align}
    \ConfMethod = \sum_{i=1}^n w_i l_i 
    \text{ where } & w_i = \frac{a_{i}}{\sum_{i'=1}^n a_{i'}} \label{eq:main}
\end{align}
where $a_{i'}$ is the attention of the last token of the attention-eliciting prompt on the $\predSeq_i$. 

\begin{figure}[ht]
\begin{lstlisting}[basicstyle=\footnotesize\ttfamily]
Read the following question with optional context and decide if the answer correctly answer the question. Focus on the answer, and reply Y or N.
...
Context: Harry is a good witcher.
Question: How old is Harry?
Answer: Harry practices witchcraft.
 Decision: N. (The answer does not mention Harry's age.)
...
[$optional_context]
Question: [$question]
Answer: [$response]
 Decision:
\end{lstlisting}
\vspace{-3mm}
\caption{The attention-eliciting prompt used in this paper (full version deferred to the Appendix due to space constraints).
{\color{blue}\$optional\_context}, {\color{blue}\$question} and {\color{blue}\$response} are replaced with the corresponding values of a sample.
In our experiments, {\color{blue}\$optional\_context} refers to the story and conversation history that accompanies each question in CoQA~\citep{reddy-etal-2019-coqa}.
}\label{fig:prompt_short}
\end{figure}

\def \FigAttentionCase{
\begin{figure*}[ht]
\begin{minipage}{0.6\textwidth}
{\small
\begin{align*}
\textbf{Q:} &\text{\colorbox{red!40}{When did Neil Armstrong land on the Moon?}}\\
&\text{\colorbox{blue!40}{Who landed on the Moon with Neil Armstrong on July 20, 1969?}}\\
&\text{\colorbox{yellow!40}{What did Neil Armstrong and Buzz Aldrin do on July 20, 1969?}}\\
\textbf{A:} &\underbrace{\text{\colorbox{red!40}{On July 20, 1969}}}_{\text{when}} \text{, Armstrong and} \underbrace{\colorbox{blue!40}{\text{Buzz Aldrin}}}_{\text{who}} \underbrace{\text{\colorbox{yellow!40}{landed on the Moon}}}_{\text{what}} \\ &\hspace{10pt}\text{ for the first time in human history.}&&
\end{align*}
}
\end{minipage}
\begin{minipage}{0.37\textwidth}
\includegraphics[width=0.95\textwidth]{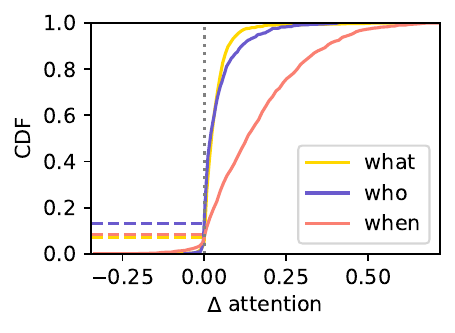} 
\end{minipage}
\vspace{-3mm}
\caption{Depending on the question, the attention-eliciting prompt introduced in \cref{fig:prompt_short} induces attention focusing on different parts of the same response (``when'', ``who'' and ``what'').
In the plot on the right, we show the CDF of $\Delta_{attn}$, the change of attention weight on the corresponding concept when asked the relevant question, on all 1,024 heads of Mistral-7B.
For example, for ``when'', we compute $\Delta_{attn}$ as the attention weight on ``On July 20, 1969'' when asked the ``when'' question minus the average of the cases where the other two questions were asked.
In all cases, the attention significantly increases on the relevant tokens (p-value from one-sided t-test is at most 9e-90).
}
\label{fig:attnillustration}
\vspace{-3mm}
\end{figure*}
}
\FigAttentionCase

\cref{fig:attnillustration} illustrates how the attention-eliciting prompt (\cref{fig:prompt_short}) induces varying emphases on the same response for different questions.
For instance, when the question focuses on ``when'', the section of the response detailing the event date receives greater attention weight across all heads. 
This indicates that the weighting scheme introduced in \cref{eq:main} effectively overweights relevant tokens and underweights irrelevant ones, resulting in a more contextualized version of sequence likelihood. 

Besides using a prompt, a more straightforward source of attention weights is the original generating process.
Specifically, when the LM completes generating $\predSeq$, we retrieve the attention for the next token.
We refer to this variant as \methodnameNext, in contrast to \methodnamePrompt with the prompt. 
Our hypothesis is that the LM induces some internal attention on the critical words of the generation even without an explicit prompt asking for it, and we will explain how to identify such attention in \cref{sec:method:headchoice}.


\subsection{The Choice of Heads}\label{sec:method:headchoice}

While the prompt can enhance overall attention on relevant tokens, averaging attention across heads---even with the prompt---is unwise. 
Many attention heads likely focus, for example, on ensuring grammatical correctness, with only a fraction dedicated to the response. 
Selecting only the useful heads from the multitude of heads (e.g., out of 1,600 in LLaMA2-13B) remains a challenging task.

We present a systematic approach to identify the appropriate heads.
Let $w^h$  denote the attention weights from head $h$, and $\ConfMethod^h$ the associated confidence score computed via  \cref{eq:main}.
For each head $h$ and a subset of the samples, we use the associated confidence $\ConfMethod^h$ to predict the accuracy of responses, and compute an AUROC (similar to \citealp{kuhn2023semantic}), denoted as $\text{AUROC}_h$. 
Notably, we found that the ``functionality'' of the heads appears relatively stable: That is, if we compare the $\text{AUROC}_h$ on two subsets of the population, the ranking of these heads is highly consistent across the two subsets, as shown in \cref{fig:headvaltest}. 
As a result, we propose to pick the top $k=10$ heads on the validation dataset and average the attention weights of only these heads.
The reason why we pick more than one head is because picking only the best head is likely affected by noise due to the size of the validation set.
In \cref{sec:exp}, we show that leveraging the attention from about $k=10$ top heads performs better than either the average attention of all heads, or the top head.

\def \FigHeadConsistency{
\begin{figure}[ht]
\centering
\includegraphics[width=0.42\textwidth]{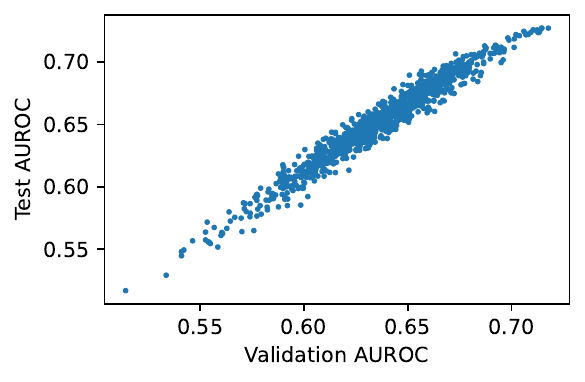} 
\vspace{-3mm}
\caption{
Scatter plot of test vs validation AUROC for confidence measures computed via \cref{eq:main} with different heads' attention weights, on Natural Questions (\datasetnqopen) 
with Mistral-7B model. 
The ranking is highly consistent---the best heads on the validation set continue to perform well on the test set. 
In this case, the Spearman correlation~\citep{spearman1961proof} is $>97\%$.
We can thus pick only a small subset of the 1024 heads (or more for other LMs) to construct the final confidence measure $\ConfMethod$.
}
\label{fig:headvaltest}
\vspace{-4mm}
\end{figure}
}
\FigHeadConsistency

\section{Experiments}\label{sec:exp}
\subsection{Datasets}

We use the following standard benchmark datasets, largely following the practices in \citet{kuhn2023semantic,lin2023generating}:
\begin{itemize}
    \item CoQA (\datasetcoqa)~\citep{reddy-etal-2019-coqa}, an open-book conversational question answering dataset. 
    We use the development split of \datasetcoqa with 7,983 questions.
    
    \item TriviaQA (\datasettrivia)~\citep{joshi-etal-2017-triviaqa}, a closed-book QA dataset.
    We use the validation split of the \texttt{rc.nocontext} subset of \datasettrivia with 9,960 (de-duplicated) questions. 

    \item Natural Questions (\datasetnqopen)~\citep{kwiatkowski-etal-2019-natural}, a closed-book QA dataset.
    We use the validation split of \datasetnqopen with 3,610 questions.
\end{itemize}
Following~\cite{lin2023generating}, for each experiment we use a random subset of 1,000 questions as the validation set, and the remaining as the test set. 
We report the mean and standard deviation of all evaluation metrics (see~\cref{sec:exp:eval}) on the test set, calculated from 10 random data splittings.

\subsection{Baselines}
We compare $\methodnamePrompt$ (and $\methodnameNext$) with several recent confidence measures:
\begin{itemize}

    \item $\baselineDegree$, a confidence score based on the degree of similarity graph from~\citet{lin2023generating}.
    Note that this method requires sampling multiple responses, and we set the number of additional generations to $5$.
    We use the ``Entailment'' version as suggested by \citet{lin2023generating}.
    
    \item \baselinePTrue~\citep{kadavath2022language}, which elicits the confidence by asking the LM itself whether its response is correct.
    We use the prompt from~\citet{kadavath2022language,lin2023generating}. 

    \item Sequence Likelihood (\baselineNLL): 
    This is the sequence likelihood measure discussed in \cref{eq:confll}, and widely in literature as a confidence score~\cite{lin2023generating,quach2023conformal,huang2024uncertainty} or as a building block for predictive entropy~\citep{kuhn2023semantic,kadavath2022language,malinin2021uncertainty}.
    We include the length-normalized version in \cref{eq:confll:norm}, \baselineNLLNorm, as well.
    
    \item \baselineSAR~\citep{duan2023shifting}:
    It proposes to estimate the relevance of each token as $w_i$ in \cref{eq:main}, with $w_i\propto 1-sim(\predSeq, \predSeq_{-i})$ where $sim$ is similarity measured by an NLI model 
    and $\predSeq_{-i}$ is the response of interest $\predSeq$ without token $i$.

\end{itemize}
In addition, we also replace the sequence likelihood used in Semantic Entropy (\baselineSE,\citealp{kuhn2023semantic}), which is an \textit{uncertainty}\footnote{See \citet{lin2023generating} for a discussion distinguishing between confidence and uncertainty.} measure, with \methodnamePrompt, and report results in \cref{sec:exp:uncertainty}

\subsection{Language Model and Generation}
For the base LLMs, we include the most popular open LLMs: LLaMA2~\citep{touvron2023llama2}, Mistral~\citep{jiang2023mistral} and Gemma~\citep{team2024gemma}.
We use the 13B version for LLaMA, and the 7B version for Mistral and Gemma due to their improved performance.
Response generation largely follows ~\citet{kuhn2023semantic} with some improvements: The original pipeline often removes content after periods in abbreviations (such as ``Dr.''), so we modified the prompt to ensure each response $\predSeq$ ends with a newline character but keeps contents after punctuations.
Like \citet{kuhn2023semantic}, we focus on the greedily-decoded generation $\predSeq$ for each question and use a temperature of $0.5$ for baselines that require additional response sampling. 


\def \TabAutoEvalAgreement{
\begin{table}[ht]
\caption{
Agreement with human annotation, on 720 sampled question-response pairs in total.
}
\label{table:main:huamneval}
\centering
\begin{small}
\begin{tabular}{l|cccc}
\toprule
 Agreement with Human (\%) & \datasetcoqa &  \datasetnqopen & \datasettrivia\\
\midrule
$\Acc_{agree}$ & 98.2 & 91.8 & 98.2\\
$\Acc_{llama2}$ & 97.0 & 86.7 & 95.0\\
$\Acc_{gpt}$ & 94.1 & 85.8 & 96.6 \\
\bottomrule
\end{tabular}
\end{small}
\end{table}
}
\TabAutoEvalAgreement
\def \TabAUROC{
{
\setlength{\tabcolsep}{3pt}
\begin{table*}
\caption{
AUROC of using confidence measures $\Conf$ to predict the accuracy of responses.
Methods not significantly different from the best are in \textbf{bold}. 
}
\label{table:main:exp:roc:mlg}
\centering
   \resizebox{1.8\columnwidth}{!}{
\begin{tabular}{c|ccccc|cc}
\toprule
& \baselineDegree(E) & \baselinePTrue & \baselineNLL & \baselineNLLNorm & \baselineSAR & \methodnamePrompt & \methodnameNext\\
\midrule
\datasettrivia(llama2) & 81.74$\pm$0.25 & 64.82$\pm$0.18 & 88.19$\pm$0.12 & 87.86$\pm$0.12 & 87.91$\pm$0.13 & \textbf{89.70$\pm$0.19} & \textbf{89.61$\pm$0.18}\\
\datasettrivia(gemma) & 84.00$\pm$0.19 & 81.82$\pm$0.18 & 88.72$\pm$0.11 & 88.11$\pm$0.08 & 88.09$\pm$0.08 & \textbf{89.71$\pm$0.14} & 89.42$\pm$0.11\\
\datasettrivia(mistral) & 81.85$\pm$0.30 & 68.78$\pm$0.28 & 88.81$\pm$0.15 & 88.64$\pm$0.14 & 88.74$\pm$0.13 & \textbf{90.76$\pm$0.18} & \textbf{90.73$\pm$0.15}\\
\datasetcoqa(llama2) & 69.93$\pm$0.57 & 53.64$\pm$0.41 & 69.50$\pm$0.40 & 72.59$\pm$0.44 & 72.78$\pm$0.47 & \textbf{73.34$\pm$0.74} & \textbf{73.36$\pm$0.57}\\
\datasetcoqa(gemma) & 70.03$\pm$0.68 & 55.99$\pm$0.42 & 70.83$\pm$0.46 & 71.96$\pm$0.57 & 72.38$\pm$0.55 & \textbf{73.30$\pm$0.57} & \textbf{73.64$\pm$0.63}\\
\datasetcoqa(mistral) & 69.84$\pm$0.52 & 52.33$\pm$0.42 & 68.97$\pm$0.38 & 70.60$\pm$0.38 & 70.87$\pm$0.41 & \textbf{71.79$\pm$0.75} & \textbf{71.91$\pm$0.63}\\
\datasetnqopen(llama2) & 71.61$\pm$0.51 & 52.51$\pm$0.47 & 66.57$\pm$0.33 & 69.48$\pm$0.50 & 70.43$\pm$0.46 & \textbf{73.73$\pm$0.49} & \textbf{73.54$\pm$0.46}\\
\datasetnqopen(gemma) & 73.32$\pm$0.68 & 63.66$\pm$0.46 & 72.09$\pm$0.65 & 75.81$\pm$0.65 & 75.88$\pm$0.68 & \textbf{77.95$\pm$0.58} & 77.17$\pm$0.65\\
\datasetnqopen(mistral) & 73.03$\pm$0.52 & 54.77$\pm$0.51 & 69.22$\pm$0.53 & 71.06$\pm$0.54 & 72.61$\pm$0.48 & \textbf{76.65$\pm$0.43} & 75.73$\pm$0.68\\
\bottomrule
\end{tabular}
}
\end{table*}
}}
\TabAUROC

\def \TabAUARC{
{
\setlength{\tabcolsep}{3pt}
\begin{table*}
\caption{
AUARC of using confidence measures $\Conf$ to predict the accuracy of responses.
Methods not significantly different from the best are in \textbf{bold}. 
}
\label{table:main:exp:arc:mlg}
\centering
   \resizebox{2\columnwidth}{!}{
\begin{tabular}{ccc|ccccc|cc}
\toprule
& Random & Upper Bound & \baselineDegree(E) & \baselinePTrue & \baselineNLL & \baselineNLLNorm & \baselineSAR & \methodnamePrompt & \methodnameNext\\
\midrule
\datasettrivia(llama2) & 82.60$\pm$0.14 & 98.39$\pm$0.03 & 92.84$\pm$0.28 & 87.50$\pm$0.12 & 95.99$\pm$0.06 & 95.95$\pm$0.05 & 95.96$\pm$0.05 & \textbf{96.32$\pm$0.08} & \textbf{96.29$\pm$0.06}\\
\datasettrivia(gemma) & 78.14$\pm$0.13 & 97.41$\pm$0.03 & 91.48$\pm$0.19 & 91.83$\pm$0.11 & 94.61$\pm$0.05 & 94.38$\pm$0.04 & 94.38$\pm$0.04 & \textbf{94.79$\pm$0.04} & 94.65$\pm$0.04\\
\datasettrivia(mistral) & 79.94$\pm$0.12 & 97.84$\pm$0.03 & 91.91$\pm$0.31 & 87.55$\pm$0.13 & 95.27$\pm$0.06 & 95.21$\pm$0.05 & 95.22$\pm$0.05 & \textbf{95.68$\pm$0.09} & \textbf{95.68$\pm$0.07}\\
\datasetcoqa(llama2) & 91.36$\pm$0.17 & 99.62$\pm$0.01 & 94.71$\pm$0.22 & 92.24$\pm$0.18 & 95.69$\pm$0.13 & 96.09$\pm$0.14 & 96.17$\pm$0.14 & \textbf{96.26$\pm$0.18} & \textbf{96.26$\pm$0.16}\\
\datasetcoqa(gemma) & 92.64$\pm$0.14 & 99.72$\pm$0.01 & 95.46$\pm$0.20 & 94.13$\pm$0.15 & 96.54$\pm$0.10 & 96.63$\pm$0.11 & 96.68$\pm$0.11 & \textbf{96.85$\pm$0.11} & \textbf{96.90$\pm$0.13}\\
\datasetcoqa(mistral) & 92.04$\pm$0.14 & 99.67$\pm$0.01 & 95.09$\pm$0.33 & 92.62$\pm$0.16 & 95.92$\pm$0.10 & 96.13$\pm$0.11 & 96.22$\pm$0.10 & \textbf{96.37$\pm$0.13} & \textbf{96.38$\pm$0.12}\\
\datasetnqopen(llama2) & 56.49$\pm$0.68 & 88.74$\pm$0.39 & 70.59$\pm$1.27 & 57.68$\pm$0.68 & 70.51$\pm$0.76 & 71.01$\pm$0.81 & 71.97$\pm$0.80 & \textbf{73.46$\pm$0.66} & \textbf{73.31$\pm$0.87}\\
\datasetnqopen(gemma) & 47.16$\pm$0.65 & 82.59$\pm$0.49 & 62.96$\pm$0.94 & 57.30$\pm$0.67 & 65.78$\pm$0.95 & 66.41$\pm$0.92 & 67.02$\pm$0.93 & \textbf{67.78$\pm$1.07} & 66.89$\pm$1.03\\
\datasetnqopen(mistral) & 52.90$\pm$0.74 & 86.57$\pm$0.47 & 67.48$\pm$1.00 & 56.47$\pm$0.93 & 69.15$\pm$0.82 & 69.27$\pm$0.72 & 70.62$\pm$0.65 & \textbf{72.25$\pm$0.74} & 71.85$\pm$0.69\\
\bottomrule
\end{tabular}
}
\end{table*}
}}
\TabAUARC

\subsection{Evaluation}\label{sec:exp:eval}

For an effective confidence measure, low confidence should correlate with a higher probability of incorrect generation.
To assess the quality of confidence measures, we adhere to established methodologies \citep{kuhn2023semantic,Band2022BenchmarkingBD}.
This involves utilizing them to predict the correctness of a generation and calculating the Area Under the Receiver Operating Characteristic curve (\textbf{AUROC}) for this prediction task.
Formally, let $\Acc_{i}$ represent the indicator function $\mathbbm{1}\{\predSeq_i \text{ correctly answers } x_i\}$. 
We compute the AUROC using the confidence measure $\Conf(x_\cdot, \predSeq_\cdot)$ to predict $\Acc_\cdot$. 
This approach allows us to systematically evaluate how well the confidence measures distinguish between correct and incorrect responses across multiple samples.

In addition to AUROC, following~\citet{lin2023generating}, we also report Area Under Accuracy-Rejection Curve (\textbf{AUARC})~\citep{auarc}.
The Accuracy-Rejection Curve (ARC) computes the average the accuracy when a subset of samples is rejected based on $\Conf$. 
As we exclude more low-confidence samples, the accuracy of the remaining samples should increase. 
The \textit{upper bound} is achieved by predicting only the correct $\predSeq$ (i.e. set $\Conf$ to accuracy), and the AUARC of a \textit{random} predictor is equal to the base accuracy without rejection.

\textbf{Correctness of Generations}:
A critical requirement for computing AUROC or AUARC is a reliable $\Acc_{i}$, the accuracy of each response. 
This is a unique challenge in UQ for NLG, and deserves separate research. 
Prior work typically relies on lexical similarity measures such as ROUGE~\citep{kuhn2023semantic,quach2023conformal} or BLEU~\citep{huang2024uncertainty}.
\citet{lin2023generating} uses \texttt{gpt-3.5} to evaluate the correctness of each response given a reference answer\footnote{The original paper uses \texttt{gpt-3.5-turbo-0301} which is no longer accessible.}, and considers anything with a rating above 70\% as correct.
Inspired by~\cite{lin2023generating}, we use the agreement of both \LLaMATwoName and \GPTName's evaluations as $\Acc$ (More details are in \cref{appendix:autoeval}).
As shown in \cref{table:main:huamneval}, this notably improves the correctness evaluation and thus the reliability of AUROC/AUARC.
We include the results using $\Acc_{llama2}$ in the \cref{appendix:llama2_acc} since $\Acc_{agree}$ and $\Acc_{gpt}$ may not remain reproducible in the future.


\subsection{Results}\label{sec:exp:result}
\cref{table:main:exp:roc:mlg} presents the AUROC of different confidence measures.
Clearly, all the confidence baselines consistently detect good $\predSeq$ over bad ones, but \methodname outperforms baselines.
Similar results are observed for AUARC in \cref{table:main:exp:arc:mlg} as well: As the LMs have good base accuracy (from the ``Random'' column) for \datasetcoqa and \datasettrivia, the gap between different confidence measures is relatively small, but generally significant.
In particular, among likelihood-based methods, it is sometimes confusing whether normalized or unnormalized likelihood should be used~\citep{kuhn2023semantic,malinin2021uncertainty}, and 
\baselineSAR does not always outperform the better of the two.
However, \methodname consistently outperforms all three.
Thanks to the additional sampling, \baselineDegree performs quite well, especially if we further increase the temperature of the base LM (see Appendix), but in practice it could significantly increase the computation cost as it requires $m$ additional generations and $O(m^2)$ similarity comparisons.
Finally, after the head selection process, the difference between \methodnamePrompt and \methodnameNext is small, but still extremely significant if we perform a pooled test. 
The observed similar performance is because chosen attention heads exhibit highly correlated patterns (see \cref{sec:exp:deeperattn}).
We recommend \methodnamePrompt over \methodnameNext as they share similar overhead (close to none) but the attention-eliciting prompt is more structured compared with the more arbitrary prompt used to generate $\predSeq$, and therefore the ``good'' heads are likely to be more stable.

\def \TabUQ{
{
\setlength{\tabcolsep}{3pt}
\begin{table}
\caption{
AUROC of using variants of Semantic Entropy to predict the accuracy of responses.
\methodSEAttn significantly improves the original version based on vanilla sequence likelihoods.
}
\label{table:main:exp:roc:uq}
\centering
   \resizebox{.9\columnwidth}{!}{
\begin{tabular}{c|cc|c}
\toprule
  & \baselineSENorm & \baselineSE & \methodSEAttn\\
\midrule
\datasettrivia(llama2) & 89.88$\pm$0.11 & 89.33$\pm$0.12 & \textbf{90.50$\pm$0.12}\\
\datasettrivia(gemma) & 90.33$\pm$0.10 & 90.02$\pm$0.12 & \textbf{90.75$\pm$0.13}\\
\datasettrivia(mistral) & 90.35$\pm$0.12 & 89.78$\pm$0.15 & \textbf{91.13$\pm$0.13}\\
\datasetcoqa(llama2) & \textbf{75.26$\pm$0.33} & 72.50$\pm$0.34 & \textbf{75.58$\pm$0.51}\\
\datasetcoqa(gemma) & \textbf{74.81$\pm$0.48} & 72.95$\pm$0.48 & \textbf{75.08$\pm$0.50}\\
\datasetcoqa(mistral) & 74.06$\pm$0.41 & 71.54$\pm$0.32 & \textbf{74.56$\pm$0.51}\\
\datasetnqopen(llama2) & 74.13$\pm$0.51 & 69.62$\pm$0.38 & \textbf{76.33$\pm$0.50}\\
\datasetnqopen(gemma) & 77.93$\pm$0.58 & 73.67$\pm$0.77 & \textbf{79.50$\pm$0.52}\\
\datasetnqopen(mistral) & 75.71$\pm$0.40 & 71.85$\pm$0.49 & \textbf{78.66$\pm$0.35}\\
\bottomrule
\end{tabular}
}
\end{table}
}}
\TabUQ
\vspace{-1mm}
\subsection{Improving Uncertainty Measures}\label{sec:exp:uncertainty}
As noted earlier, sequence likelihood is widely used in entropy computation, which is used as an uncertainty measure for NLG.
Semantic Entropy~\citep{kuhn2023semantic} is a state-of-the-art uncertainty measure that groups sampled generations into semantic sets and computes the entropy over these sets.
In doing so, it uses sequence likelihood (sometimes normalized). We simply replace it with \methodnamePrompt to create a new uncertainty measure, \methodSEAttn.
The comparison is shown in \cref{table:main:exp:roc:uq}.
\methodSEAttn consistently outperforms either the normalized or the unnormalized version of $\baselineSE$, showing potential in replacing sequence likelihood in other domains such as conformal NLG~\cite{quach2023conformal}.

\subsection{Is the Improvement a Fluke?}\label{sec:exp:deeperattn}

\def \FigAttnCorr{
\begin{figure}[ht]
\centering
\includegraphics[width=0.45\textwidth]{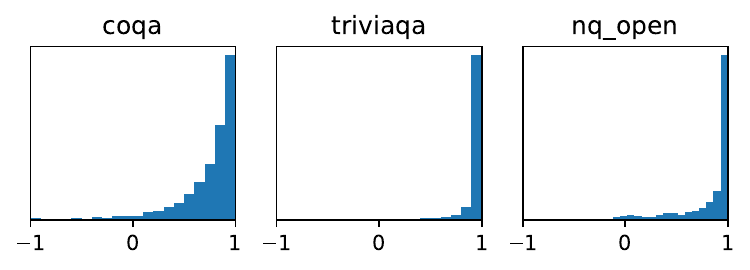}
\caption{
Histogram of the correlation between attentions from \methodnamePrompt and \methodnameNext (top 10 heads' average).
We keep only generations with more than 2 tokens.
For most responses, the chosen heads' attentions are highly correlated, suggesting that both methods focus on the same tokens, as exemplified in \cref{fig:main:case}.
}
\label{fig:attncorr}
\end{figure}
}
\FigAttnCorr

Despite not using an explicit attention-eliciting prompt, \methodnameNext performs close to \methodnamePrompt, outperforming baselines significantly.
It is reasonable to then worry that the improvement is just the result of black-box data-mining by the head-selection step in \cref{sec:method:headchoice}.
This section presents evidence that \methodname most likely identifies meaningful concepts.

\cref{fig:attncorr} shows the correlation between the $w_i$ vectors used for \methodnamePrompt and \methodnameNext---which is almost always positive and usually close to 1. 
As the prompts used to induce these attention weights are quite different, such agreement could be taken as preliminary evidence that these weights are ``meaningful'' and less likely to be purely the results of two independent black-box ``fitting'' processes.

\begin{figure}[ht]
\begin{small}
  \begin{framed}
\textit{Question 1:} how early did he want to get there?\\
\textit{Response 1:} \attnplus{an hour before} the time\\
\textit{Question 2:} What does Barwell think of him?\\
\textit{Response 2:} \attnplus{he} is \attnplus{not fit} to be his guard\attnplus{ian}\\
\textit{Question 3:} who is Susan Boyle?\\
\textit{Response 3:} \attnplus{Susan} Bo\attnplus{yle is a Scottish singer who} became \attnplus{famous} after appearing on the TV show "Britain\attnplus{'}s Got Talent" in \attnplus{2}009\attnplus{.}
  \end{framed}
  \end{small}
\caption{
Tokens whose attention is increased are \attnplus{marked} (others decreased).
As expected, such re-weighting are not always interpretable, but help locating the more relevant tokens in general.
}\label{fig:main:case}
\end{figure}
\cref{fig:main:case} shows a few examples of the induced attention weights, illustrating how \methodname makes the confidence measure more focused on important tokens.
Given the nature of soft attention, despite the fact that $w_i$ chosen by \cref{sec:method} generally identifies the important tokens and improves upon vanilla sequence likelihood, it is still sometimes difficult to interpret why each token is under/over-weighted. 
For interpretability reasons, a future research direction might be to directly use natural language to identify such tokens. For example, one might directly ask the LM to list the important entities with respect to a question.
The challenge, then, transfers to identifying the actual tokens implied by the natural language output listing important entities in the original $\predSeq$.

Finally, if the attention weights are actually focusing on the important concepts as intended, one might expect that they should transfer \textit{between datasets} and be more robust to distribution shifts.
In \cref{table:main:exp:roc:coqa_transfer}, we select heads based on validation sets from \datasetcoqa and apply them on the other two datasets, which are quite different from \datasetcoqa (e.g. in the format of questions).
\methodnamePrompt still provides consistent performance boost, while \methodnameNext sometimes lags behind \baselineNLLNorm.
This suggests both the prompt and the head selection step increase the weights on the more relevant tokens.

\def \TabCoqaTransfer{
{
\setlength{\tabcolsep}{3pt}
\begin{table}
\caption{
AUROC using heads picked from \datasetcoqa.
Despite the big distribution shift from \datasetcoqa to \datasetnqopen/\datasettrivia, the heads chosen on \datasetcoqa still provides attention weights that significantly improves the AUROC.
}
\label{table:main:exp:roc:coqa_transfer}
\centering
   \resizebox{.9\columnwidth}{!}{
\begin{tabular}{c|ccc}
\toprule
 & \baselineNLLNorm & \methodnamePrompt & \methodnameNext\\
\midrule
\datasettrivia(llama2) & 87.86$\pm$0.12 & \textbf{88.59$\pm$0.85} & 88.37$\pm$0.87\\
\datasettrivia(gemma) & 88.11$\pm$0.08 & \textbf{89.03$\pm$0.61} & 87.43$\pm$0.94\\
\datasettrivia(mistral) & 88.64$\pm$0.14 & \textbf{89.70$\pm$1.43} & 88.46$\pm$1.93\\
\datasetnqopen(llama2) & 69.48$\pm$0.50 & \textbf{70.94$\pm$1.66} & 70.10$\pm$2.09\\
\datasetnqopen(gemma) & 75.81$\pm$0.65 & \textbf{77.21$\pm$1.06} & 74.99$\pm$1.30\\
\datasetnqopen(mistral) & \textbf{71.06$\pm$0.54} & \textbf{72.71$\pm$3.32} & \textbf{72.27$\pm$2.72}\\
\bottomrule
\end{tabular}
}
\end{table}
}}
\TabCoqaTransfer

\subsection{Ablation: Choice of Attention Heads}
\def \TabHeadChoice{
{
\setlength{\tabcolsep}{3pt}
\begin{table}
\caption{
AUROC of different choice of heads.
Using all heads, heads from last layer, or the best head are generally better than equal-weighted (\baselineNLLNorm) but worse than using the best 10 heads (\methodname).
}
\label{table:main:exp:roc:head_choice}
\centering
   \resizebox{1\columnwidth}{!}{
\begin{tabular}{c|cccc|c}
\toprule
  & \baselineNLLNorm & \methodname(best) & \methodname(all) & \methodname(last) & \methodname ($k=10$)\\
\midrule
triviaqa(llama2) & 87.86$\pm$0.12 & 89.26$\pm$0.48 & 87.75$\pm$0.12 & 87.17$\pm$0.12 & \textbf{89.70$\pm$0.19}\\
triviaqa(gemma) & 88.11$\pm$0.08 & 89.32$\pm$0.31 & 88.08$\pm$0.08 & 88.35$\pm$0.07 & \textbf{89.71$\pm$0.14}\\
triviaqa(mistral) & 88.64$\pm$0.14 & 90.46$\pm$0.27 & 89.18$\pm$0.12 & 88.41$\pm$0.14 & \textbf{90.76$\pm$0.18}\\
coqa(llama2) & 72.59$\pm$0.44 & 72.46$\pm$1.25 & \textbf{73.14$\pm$0.49} & \textbf{73.09$\pm$0.49} & \textbf{73.34$\pm$0.74}\\
coqa(gemma) & 71.96$\pm$0.57 & 72.34$\pm$1.00 & 72.79$\pm$0.59 & 72.01$\pm$0.58 & \textbf{73.30$\pm$0.57}\\
coqa(mistral) & 70.60$\pm$0.38 & 70.86$\pm$1.07 & \textbf{71.35$\pm$0.38} & \textbf{71.66$\pm$0.38} & \textbf{71.79$\pm$0.75}\\
nq(llama2) & 69.48$\pm$0.50 & \textbf{73.23$\pm$0.77} & 68.72$\pm$0.51 & 67.70$\pm$0.55 & \textbf{73.73$\pm$0.49}\\
nq(gemma) & 75.81$\pm$0.65 & \textbf{77.45$\pm$0.89} & 76.37$\pm$0.65 & 76.76$\pm$0.69 & \textbf{77.95$\pm$0.58}\\
nq(mistral) & 71.06$\pm$0.54 & \textbf{76.41$\pm$0.50} & 70.83$\pm$0.57 & 69.73$\pm$0.60 & \textbf{76.65$\pm$0.43}\\
\bottomrule
\end{tabular}
}
\end{table}
}}

\def \FigHeadChoice{
\begin{figure}[ht]
\centering
\includegraphics[width=0.48\textwidth]{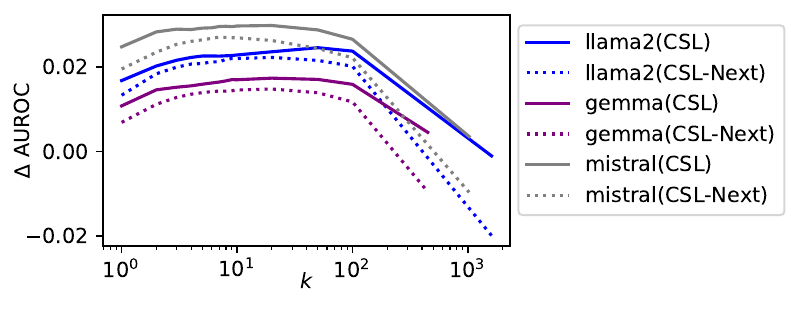}
\vspace{-7mm}
\caption{
AUROC gain compared with \baselineNLLNorm for different $k$ (number of attention heads to keep), from 1 to all heads.
The performance peaks around $k=10$ and is stable.
\methodnamePrompt is also consistently better than \methodnameNext - notably, when we average the attention from all heads, \methodnamePrompt still outperforms \baselineNLLNorm, but \methodnameNext is significantly worse.
}
\label{fig:headchoice}
\vspace{-3mm}
\end{figure}
}
\FigHeadChoice

In \cref{fig:headchoice}, we compute the \textit{gain} in AUROC compared with \baselineNLLNorm (i.e. equal weighting), keeping different number of attention heads.
The solid lines represent \methodnamePrompt, and the dotted lines denote \methodnameNext.
Note that using only one head is also significantly better than the baseline, but performance increases and peaks around 10 heads (sometimes more). 
We believe using a small number of ``good'' heads can reduce the noise introduced by using a small validation set, making \methodname more contextualized to the question.

\section{Discussion and Conclusion}\label{sec:conclusion}
In this paper, we explore enhancements to the widely used confidence measure, sequence likelihood, which serves as a quality metric for model generations in selective generation and risk control for natural language generation. 
We introduce Contextualized Sequence Likelihood, or \methodname, a novel approach that utilizes attention weights on generated tokens to re-weight the logits in sequence likelihood computation. 
This new confidence measure surpasses existing methods across several popular datasets and large language models by more accurately predicting the accuracy of each response.

Despite these improvements, there are limitations to the current approach. 
First, the interpretability of attention weights is often obscured by the nature of the self-attention mechanism. 
While attention re-weighting generally enhances the confidence measure, there are individual cases where selected heads do not align with tokens that humans would typically consider important in the context of the question. 
As discussed in \cref{sec:exp}, a possible solution involves identifying key tokens through a language model via natural language, though this introduces the additional challenge of matching these words to the original response.

Another limitation is the applicability of the current prompt, which is tailored for question-answering and may require modifications for other tasks.
Additionally, like many baseline methods, the current approach cannot leverage external information for ``fact-checking.'' 
Integrating confidence from multiple models could potentially bridge the gap between a language model's perceived confidence and the actual correctness of a response.
We hope that future research will address these issues and expand the toolkit available to practitioners for assessing the reliability of large language models.



\clearpage 
\bibliography{main}

\appendix
\section{Prompts}\label{appendix:prompt}
In this section, we present the full prompts used in various aspects of our experiments.

\subsection{Question-Answering Generation Prompts}
We use the following prompts when generating responses for the question-answering datasets.

\textbf{CoQA}:
\begin{lstlisting}[basicstyle=\footnotesize\ttfamily]
Read the context and answer the questions below.

*Context*: [$context]
[additional question-answer pairs]
*Question*: [$question]
*Answer*:
\end{lstlisting}
where additional question-answer pairs are preceding turns of the conversation about the paragraph consisting of questions and reference answers.

\textbf{TriviaQA}:
\begin{lstlisting}[basicstyle=\footnotesize\ttfamily]
Answer these questions:

*Question*: In Scotland a bothy/bothie is a?
*Answer*: House
*Question*: [$question]
*Answer*:
\end{lstlisting}

\textbf{Natural Questions} is a much harder dataset than TriviaQA, so we use the same 5-shot prompt version of the prompt in~\cite{touvron2023llama} (with 5 questions randomly picked from the training set).
\begin{lstlisting}[basicstyle=\footnotesize\ttfamily]
Answer these questions:

*Question*: who makes up the state council in russia
*Answer*: governors and presidents
*Question*: when does real time with bill maher come back
*Answer*: November 9, 2018
*Question*: where did the phrase american dream come from
*Answer*: the mystique regarding frontier life
*Question*: what do you call a group of eels
*Answer*: bed
*Question*: who wrote the score for mission impossible fallout
*Answer*: Lorne Balfe
*Question*: [$question]
*Answer*:
\end{lstlisting}

\subsection{Attention-eliciting Prompt}
In the following, we provide the full prompt previewed in \cref{fig:prompt_short}.
\begin{lstlisting}[basicstyle=\footnotesize\ttfamily]
Read the following question with optional context and decide if the answer correctly answer the question. Focus on the answer, and reply Y or N.


Context: Luxor International Airport is a airport near Luxor in Egypt (EG). It is 353km away from the nearest seaport (Duba). The offical IATA for this airport is LXR.
Question: Luxor international airport is in which country?
Answer: It is in the United States, and its IATA is LXR.
 Decision: N. (The airport is in Egypt, not the United States.)


Context: Harry is a good witcher.
Question: How old is Harry?
Answer: Harry practices witchcraft.
 Decision: N. (The answer does not mention Harry's age.)


Question: What is the capital of Kenya?
Answer: Nairobi is the capital of Kenya.
 Decision: Y.


Question: Who has won the most Premier League titles since 2015?
Answer: Manchester City have win the most Premier League title after 2015.
 Decision: Y. (Grammar errors are ignored.)


[$optional_context]
Question: [$question]
Answer: [$response]
 Decision:
\end{lstlisting}

\subsection{Automatic Accuracy Evaluation}
For the prompts used to elicit judgment from \GPTName and \LLaMATwoName, we use the same ones from the Appendix of~\citet{lin2023generating}.

\section{Automatic Accuracy Evaluation}\label{appendix:autoeval}

To verify the efficacy of automatic accuracy evaluation by \GPTName and \LLaMATwoName, we compare their judgements' alignment with human annotations.
Specifically, we first sample 80 (question, response) pairs for each (model, dataset), resulting in 720 samples in total.
We then manually compare each sample's response with the reference answer, and decide if the generated response is correct (given the context if any).
Then, we retrieve the ratings on these 720 samples from \GPTName and \LLaMATwoName, and find the thresholds that result in the highest agreement.
The resulting accuracy-threshold relation is illustrated in \cref{fig:appendix:autoeval_thres}.
From the results, we chose 0.2 as the threshold for \LLaMATwoName and 0.6 for \GPTName.
In other words, denote the rating from \LLaMATwoName on a generated response $\predSeq$ as $r_{llama2}(\predSeq)$, then $\text{acc}_{llama2}(\predSeq) = \mathbbm{1}\{r_{llama2}(\predSeq) \geq 0.2\}$.
Note that we found 20 out of the 720 samples hard to decide during our manual annotation process, usually due to incorrect reference answers or intrinsic ambiguity in the questions. 
We ignore them during the computation of \cref{fig:appendix:autoeval_thres}.

\def \FigAppendixAEThres{
\begin{figure}[ht]
\centering
\includegraphics[width=0.4\textwidth]{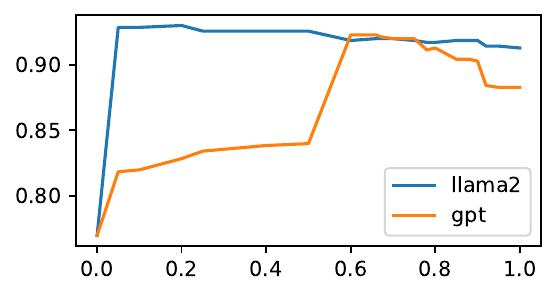}
\caption{
Agreement with human annotation with different thresholds over all 720 samples.
}
\label{fig:appendix:autoeval_thres}
\end{figure}
}
\FigAppendixAEThres

\section{Temperature of Base Language Model}\label{appendix:temperature}

We include results using a high temperature of 1.0 in \cref{table:appendix:exp:roc:mlg:t1,table:appendix:exp:arc:mlg:t1}.
Higher temperature increases the sampling diversity of \baselineDegree and \baselinePTrue, which increases the performance of \baselineDegree significantly, but brings mixed results to \baselinePTrue.
We also note that with 5 additional generations (6 in total) \baselineDegree is sometimes better than \methodnamePrompt or \methodnameNext.
We repeat the uncertainty experiments with higher temperature as well, in \cref{table:appendix:exp:roc:uq:t1}, and found that \methodSEAttn is the most predictive of the generations' quality at this temperature.

The experiments here suggest that sampling from an appropriate temperature really helps quantifying the reliability of the generations, but it should be noted that either sampling based confidence measures (\baselineDegree,\baselinePTrue) or sampling-based uncertainty measures (\baselineSENorm,\baselineSE,\methodSEAttn) bear a significantly higher computational overhead - $(m+1)\times$ the generation cost and up to $(m^2-m)$ pairwise NLI inference, with $m$ being the additional generations.
On the other hand, \methodname requires one inference (not generation) call to the base LM which is only a fraction of the overhead.
In practice, one might choose the best practice basing on cost and latency tolerance as well as difference in performance for the particular data distribution, and as shown in \methodSEAttn, \methodname could be combined with sampling to yield better results as well.

\def \TabAppendixAUROCTemp{
{
\setlength{\tabcolsep}{3pt}
\begin{table*}
\caption{
Similar to \cref{table:main:exp:roc:mlg}, but with different number of generations at temperature of 1.0.
}
\label{table:appendix:exp:roc:mlg:t1}
\centering
   \resizebox{2\columnwidth}{!}{
\begin{tabular}{c|ccccc|cc}
\toprule
 & \baselineDegree(E) & \baselinePTrue & \baselineNLL & \baselineNLLNorm & \baselineSAR & \methodnamePrompt & \methodnameNext\\
\midrule
3 generations \\ 
\midrule
\datasettrivia(llama2) & 86.49$\pm$0.14 & 61.63$\pm$0.27 & 88.16$\pm$0.13 & 87.84$\pm$0.13 & 87.89$\pm$0.13 & \textbf{89.65$\pm$0.20} & \textbf{89.58$\pm$0.17}\\
\datasettrivia(gemma) & 87.00$\pm$0.12 & 77.87$\pm$0.17 & 88.83$\pm$0.11 & 88.10$\pm$0.09 & 88.07$\pm$0.09 & \textbf{89.73$\pm$0.16} & 89.43$\pm$0.10\\
\datasettrivia(mistral) & 87.01$\pm$0.21 & 73.75$\pm$0.16 & 88.85$\pm$0.15 & 88.66$\pm$0.14 & 88.75$\pm$0.13 & \textbf{90.77$\pm$0.17} & \textbf{90.75$\pm$0.13}\\
\datasetcoqa(llama2) & 72.51$\pm$0.29 & 53.83$\pm$0.40 & 69.48$\pm$0.39 & 72.67$\pm$0.43 & \textbf{72.88$\pm$0.46} & \textbf{73.46$\pm$0.69} & \textbf{73.45$\pm$0.56}\\
\datasetcoqa(gemma) & \textbf{73.27$\pm$0.44} & 56.82$\pm$0.57 & 70.78$\pm$0.47 & 72.07$\pm$0.52 & 72.46$\pm$0.49 & 73.07$\pm$0.52 & \textbf{73.60$\pm$0.52}\\
\datasetcoqa(mistral) & \textbf{71.86$\pm$0.57} & 55.04$\pm$0.47 & 68.98$\pm$0.38 & 70.61$\pm$0.38 & 70.89$\pm$0.41 & \textbf{71.79$\pm$0.74} & \textbf{71.91$\pm$0.63}\\
\datasetnqopen(llama2) & 72.47$\pm$0.50 & 50.33$\pm$0.55 & 66.64$\pm$0.35 & 69.48$\pm$0.50 & 70.41$\pm$0.46 & \textbf{73.83$\pm$0.52} & \textbf{73.59$\pm$0.48}\\
\datasetnqopen(gemma) & 75.78$\pm$0.68 & 62.57$\pm$0.41 & 72.09$\pm$0.65 & 75.81$\pm$0.65 & 75.88$\pm$0.68 & \textbf{77.96$\pm$0.57} & 77.19$\pm$0.63\\
\datasetnqopen(mistral) & 73.76$\pm$0.70 & 58.65$\pm$0.51 & 69.22$\pm$0.53 & 71.06$\pm$0.54 & 72.61$\pm$0.48 & \textbf{76.65$\pm$0.43} & 75.73$\pm$0.68\\
\midrule
5 generations \\ 
\midrule
\datasettrivia(llama2) & 88.79$\pm$0.12 & 61.63$\pm$0.27 & 88.16$\pm$0.13 & 87.84$\pm$0.13 & 87.89$\pm$0.13 & \textbf{89.65$\pm$0.20} & \textbf{89.58$\pm$0.17}\\
\datasettrivia(gemma) & 89.19$\pm$0.11 & 77.87$\pm$0.17 & 88.83$\pm$0.11 & 88.10$\pm$0.09 & 88.07$\pm$0.09 & \textbf{89.73$\pm$0.16} & 89.43$\pm$0.10\\
\datasettrivia(mistral) & 89.39$\pm$0.21 & 73.75$\pm$0.16 & 88.85$\pm$0.15 & 88.66$\pm$0.14 & 88.75$\pm$0.13 & \textbf{90.77$\pm$0.17} & \textbf{90.75$\pm$0.13}\\
\datasetcoqa(llama2) & \textbf{75.73$\pm$0.30} & 53.83$\pm$0.40 & 69.48$\pm$0.39 & 72.67$\pm$0.43 & 72.88$\pm$0.46 & 73.46$\pm$0.69 & 73.45$\pm$0.56\\
\datasetcoqa(gemma) & \textbf{76.40$\pm$0.50} & 56.82$\pm$0.57 & 70.78$\pm$0.47 & 72.07$\pm$0.52 & 72.46$\pm$0.49 & 73.07$\pm$0.52 & 73.60$\pm$0.52\\
\datasetcoqa(mistral) & \textbf{74.41$\pm$0.51} & 55.04$\pm$0.47 & 68.98$\pm$0.38 & 70.61$\pm$0.38 & 70.89$\pm$0.41 & 71.79$\pm$0.74 & 71.91$\pm$0.63\\
\datasetnqopen(llama2) & \textbf{74.09$\pm$0.53} & 50.33$\pm$0.55 & 66.64$\pm$0.35 & 69.48$\pm$0.50 & 70.41$\pm$0.46 & \textbf{73.83$\pm$0.52} & 73.59$\pm$0.48\\
\datasetnqopen(gemma) & 77.40$\pm$0.74 & 62.57$\pm$0.41 & 72.09$\pm$0.65 & 75.81$\pm$0.65 & 75.88$\pm$0.68 & \textbf{77.96$\pm$0.57} & 77.19$\pm$0.63\\
\datasetnqopen(mistral) & \textbf{76.31$\pm$0.65} & 58.65$\pm$0.51 & 69.22$\pm$0.53 & 71.06$\pm$0.54 & 72.61$\pm$0.48 & \textbf{76.65$\pm$0.43} & 75.73$\pm$0.68\\
\bottomrule
\end{tabular}
}
\end{table*}
}}
\TabAppendixAUROCTemp

\def \TabAppendixAUARCTemp{
{
\setlength{\tabcolsep}{3pt}
\begin{table*}
\caption{
Similar to \cref{table:main:exp:arc:mlg}, but with different number of generations at temperature of 1.0.
}
\label{table:appendix:exp:arc:mlg:t1}
\centering
   \resizebox{2\columnwidth}{!}{
\begin{tabular}{ccc|ccccc|cc}
\toprule
  & Random & Upper Bound & \baselineDegree(E) & \baselinePTrue & \baselineNLL & \baselineNLLNorm & \baselineSAR & \methodnamePrompt & \methodnameNext\\
\midrule
3 generations \\ 
\midrule
\datasettrivia(llama2) & 82.61$\pm$0.14 & 98.39$\pm$0.03 & 95.04$\pm$0.18 & 87.06$\pm$0.16 & 95.99$\pm$0.06 & 95.95$\pm$0.05 & 95.96$\pm$0.05 & \textbf{96.33$\pm$0.08} & \textbf{96.30$\pm$0.07}\\
\datasettrivia(gemma) & 78.11$\pm$0.12 & 97.41$\pm$0.03 & 93.52$\pm$0.12 & 91.01$\pm$0.11 & 94.62$\pm$0.05 & 94.37$\pm$0.04 & 94.37$\pm$0.04 & \textbf{94.79$\pm$0.04} & 94.66$\pm$0.04\\
\datasettrivia(mistral) & 79.90$\pm$0.12 & 97.83$\pm$0.03 & 94.12$\pm$0.16 & 90.17$\pm$0.10 & 95.27$\pm$0.06 & 95.20$\pm$0.05 & 95.21$\pm$0.05 & \textbf{95.67$\pm$0.09} & \textbf{95.67$\pm$0.07}\\
\datasetcoqa(llama2) & 91.36$\pm$0.17 & 99.62$\pm$0.02 & 95.52$\pm$0.20 & 92.20$\pm$0.21 & 95.69$\pm$0.13 & 96.11$\pm$0.14 & \textbf{96.19$\pm$0.14} & \textbf{96.28$\pm$0.17} & \textbf{96.27$\pm$0.15}\\
\datasetcoqa(gemma) & 92.63$\pm$0.14 & 99.72$\pm$0.01 & 96.14$\pm$0.19 & 94.33$\pm$0.17 & 96.53$\pm$0.11 & 96.66$\pm$0.10 & 96.70$\pm$0.10 & 96.82$\pm$0.09 & \textbf{96.90$\pm$0.11}\\
\datasetcoqa(mistral) & 92.04$\pm$0.14 & 99.67$\pm$0.01 & 95.80$\pm$0.26 & 93.06$\pm$0.19 & 95.92$\pm$0.10 & 96.13$\pm$0.11 & 96.22$\pm$0.10 & \textbf{96.37$\pm$0.14} & \textbf{96.39$\pm$0.11}\\
\datasetnqopen(llama2) & 56.48$\pm$0.68 & 88.74$\pm$0.39 & 72.37$\pm$0.89 & 56.84$\pm$0.70 & 70.50$\pm$0.77 & 71.00$\pm$0.81 & 71.95$\pm$0.80 & \textbf{73.49$\pm$0.67} & \textbf{73.28$\pm$0.86}\\
\datasetnqopen(gemma) & 47.16$\pm$0.65 & 82.59$\pm$0.49 & 66.59$\pm$0.87 & 56.54$\pm$0.78 & 65.78$\pm$0.95 & 66.41$\pm$0.92 & 67.02$\pm$0.93 & \textbf{67.78$\pm$1.07} & 66.89$\pm$1.03\\
\datasetnqopen(mistral) & 52.90$\pm$0.74 & 86.57$\pm$0.47 & 70.09$\pm$0.84 & 59.77$\pm$0.52 & 69.15$\pm$0.82 & 69.27$\pm$0.72 & 70.62$\pm$0.65 & \textbf{72.24$\pm$0.74} & 71.83$\pm$0.67\\
\midrule
5 generations \\ 
\midrule
\datasettrivia(llama2) & 82.61$\pm$0.14 & 98.39$\pm$0.03 & 95.80$\pm$0.09 & 87.06$\pm$0.16 & 95.99$\pm$0.06 & 95.95$\pm$0.05 & 95.96$\pm$0.05 & \textbf{96.33$\pm$0.08} & \textbf{96.30$\pm$0.07}\\
\datasettrivia(gemma) & 78.11$\pm$0.12 & 97.41$\pm$0.03 & 94.51$\pm$0.08 & 91.01$\pm$0.11 & 94.62$\pm$0.05 & 94.37$\pm$0.04 & 94.37$\pm$0.04 & \textbf{94.79$\pm$0.04} & 94.66$\pm$0.04\\
\datasettrivia(mistral) & 79.90$\pm$0.12 & 97.83$\pm$0.03 & 95.15$\pm$0.11 & 90.17$\pm$0.10 & 95.27$\pm$0.06 & 95.20$\pm$0.05 & 95.21$\pm$0.05 & \textbf{95.67$\pm$0.09} & \textbf{95.67$\pm$0.07}\\
\datasetcoqa(llama2) & 91.36$\pm$0.17 & 99.62$\pm$0.02 & 96.13$\pm$0.15 & 92.20$\pm$0.21 & 95.69$\pm$0.13 & 96.11$\pm$0.14 & \textbf{96.19$\pm$0.14} & \textbf{96.28$\pm$0.17} & \textbf{96.27$\pm$0.15}\\
\datasetcoqa(gemma) & 92.63$\pm$0.14 & 99.72$\pm$0.01 & 96.74$\pm$0.20 & 94.33$\pm$0.17 & 96.53$\pm$0.11 & 96.66$\pm$0.10 & 96.70$\pm$0.10 & 96.82$\pm$0.09 & \textbf{96.90$\pm$0.11}\\
\datasetcoqa(mistral) & 92.04$\pm$0.14 & 99.67$\pm$0.01 & 96.22$\pm$0.16 & 93.06$\pm$0.19 & 95.92$\pm$0.10 & 96.13$\pm$0.11 & 96.22$\pm$0.10 & \textbf{96.37$\pm$0.14} & \textbf{96.39$\pm$0.11}\\
\datasetnqopen(llama2) & 56.48$\pm$0.68 & 88.74$\pm$0.39 & \textbf{73.77$\pm$0.89} & 56.84$\pm$0.70 & 70.50$\pm$0.77 & 71.00$\pm$0.81 & 71.95$\pm$0.80 & \textbf{73.49$\pm$0.67} & 73.28$\pm$0.86\\
\datasetnqopen(gemma) & 47.16$\pm$0.65 & 82.59$\pm$0.49 & \textbf{67.47$\pm$0.98} & 56.54$\pm$0.78 & 65.78$\pm$0.95 & 66.41$\pm$0.92 & 67.02$\pm$0.93 & \textbf{67.78$\pm$1.07} & 66.89$\pm$1.03\\
\datasetnqopen(mistral) & 52.90$\pm$0.74 & 86.57$\pm$0.47 & \textbf{72.07$\pm$0.84} & 59.77$\pm$0.52 & 69.15$\pm$0.82 & 69.27$\pm$0.72 & 70.62$\pm$0.65 & \textbf{72.24$\pm$0.74} & 71.83$\pm$0.67\\
\bottomrule
\end{tabular}
}
\end{table*}
}}
\TabAppendixAUARCTemp

{
\setlength{\tabcolsep}{3pt}
\begin{table}
\caption{
Similar to \cref{table:main:exp:roc:uq}, but with different number of generations at temperature of 1.0.
}
\label{table:appendix:exp:roc:uq:t1}
\centering
   \resizebox{1\columnwidth}{!}{
\begin{tabular}{c|cc|c}
\toprule
 & \baselineSENorm & \baselineSE & \methodSEAttn\\
\midrule
3 generations \\ 
\midrule
\datasettrivia(llama2) & 88.18$\pm$0.11 & 87.90$\pm$0.08 & \textbf{88.99$\pm$0.13}\\
\datasettrivia(gemma) & 88.05$\pm$0.13 & 88.08$\pm$0.15 & \textbf{88.73$\pm$0.15}\\
\datasettrivia(mistral) & 88.90$\pm$0.16 & 88.50$\pm$0.18 & \textbf{89.84$\pm$0.16}\\
\datasetcoqa(llama2) & \textbf{75.26$\pm$0.46} & 72.11$\pm$0.32 & \textbf{75.27$\pm$0.57}\\
\datasetcoqa(gemma) & \textbf{75.42$\pm$0.38} & 72.81$\pm$0.36 & \textbf{75.25$\pm$0.42}\\
\datasetcoqa(mistral) & 73.53$\pm$0.35 & 70.49$\pm$0.37 & \textbf{73.89$\pm$0.50}\\
\datasetnqopen(llama2) & 72.91$\pm$0.42 & 68.44$\pm$0.34 & \textbf{74.05$\pm$0.47}\\
\datasetnqopen(gemma) & 77.36$\pm$0.73 & 72.73$\pm$0.80 & \textbf{78.33$\pm$0.69}\\
\datasetnqopen(mistral) & 75.65$\pm$0.66 & 71.15$\pm$0.72 & \textbf{77.37$\pm$0.59}\\
\midrule
5 generations \\ 
\midrule
\datasettrivia(llama2) & 89.40$\pm$0.11 & 89.17$\pm$0.10 & \textbf{89.95$\pm$0.11}\\
\datasettrivia(gemma) & 89.32$\pm$0.09 & 89.31$\pm$0.12 & \textbf{89.82$\pm$0.12}\\
\datasettrivia(mistral) & 90.38$\pm$0.13 & 89.86$\pm$0.16 & \textbf{90.91$\pm$0.14}\\
\datasetcoqa(llama2) & \textbf{77.08$\pm$0.40} & 73.87$\pm$0.33 & \textbf{77.25$\pm$0.47}\\
\datasetcoqa(gemma) & \textbf{77.24$\pm$0.51} & 74.39$\pm$0.46 & \textbf{77.11$\pm$0.46}\\
\datasetcoqa(mistral) & 75.65$\pm$0.31 & 72.58$\pm$0.29 & \textbf{76.04$\pm$0.50}\\
\datasetnqopen(llama2) & 74.45$\pm$0.37 & 70.33$\pm$0.35 & \textbf{75.90$\pm$0.41}\\
\datasetnqopen(gemma) & 78.31$\pm$0.54 & 74.17$\pm$0.69 & \textbf{79.33$\pm$0.57}\\
\datasetnqopen(mistral) & 77.02$\pm$0.68 & 73.17$\pm$0.65 & \textbf{78.82$\pm$0.55}\\
\bottomrule
\end{tabular}
}
\end{table}
}
\section{Results Using $\text{acc}_{llama2}$}\label{appendix:llama2_acc}

We include results using $\text{acc}_{llama2}$ (accuracy as judged by \LLaMATwoName) for reproducibility purposes in \cref{table:appendix:exp:roc:mlg:llama2,table:appendix:exp:arc:mlg:llama2}.
Conclusions stay the same as in the main text.

{
\setlength{\tabcolsep}{3pt}
\begin{table*}
\caption{
Like \cref{table:main:exp:roc:mlg}, but using accuracy from \LLaMATwoName.
}
\label{table:appendix:exp:roc:mlg:llama2}
\centering
   \resizebox{2\columnwidth}{!}{
\begin{tabular}{c|ccccc|cc}
\toprule
  & \baselineDegree(E) & \baselinePTrue & \baselineNLL & \baselineNLLNorm & \baselineSAR & \methodnamePrompt & \methodnameNext\\
\midrule
\datasettrivia(llama2) & 80.30$\pm$0.30 & 63.90$\pm$0.17 & 86.33$\pm$0.11 & 85.83$\pm$0.12 & 85.86$\pm$0.13 & \textbf{87.72$\pm$0.24} & \textbf{87.61$\pm$0.22}\\
\datasettrivia(gemma) & 81.92$\pm$0.16 & 79.94$\pm$0.18 & 86.21$\pm$0.10 & 85.56$\pm$0.09 & 85.51$\pm$0.08 & \textbf{87.14$\pm$0.14} & 86.89$\pm$0.11\\
\datasettrivia(mistral) & 79.95$\pm$0.25 & 67.67$\pm$0.27 & 86.33$\pm$0.13 & 86.14$\pm$0.13 & 86.23$\pm$0.11 & \textbf{88.31$\pm$0.18} & \textbf{88.22$\pm$0.13}\\
\datasetcoqa(llama2) & 68.32$\pm$0.62 & 53.41$\pm$0.41 & 68.17$\pm$0.33 & 71.04$\pm$0.53 & 71.29$\pm$0.54 & \textbf{71.93$\pm$0.68} & \textbf{71.70$\pm$0.63}\\
\datasetcoqa(gemma) & 69.19$\pm$0.62 & 54.90$\pm$0.36 & 69.67$\pm$0.42 & 70.39$\pm$0.63 & 70.73$\pm$0.59 & \textbf{71.65$\pm$0.63} & \textbf{71.96$\pm$0.64}\\
\datasetcoqa(mistral) & 68.52$\pm$0.52 & 52.66$\pm$0.37 & 67.72$\pm$0.30 & 69.07$\pm$0.39 & 69.30$\pm$0.40 & \textbf{70.31$\pm$0.79} & \textbf{70.18$\pm$0.66}\\
\datasetnqopen(llama2) & 68.69$\pm$0.42 & 51.70$\pm$0.38 & 64.47$\pm$0.44 & 66.07$\pm$0.42 & 66.64$\pm$0.36 & \textbf{69.86$\pm$0.52} & 69.53$\pm$0.45\\
\datasetnqopen(gemma) & 69.73$\pm$0.57 & 60.51$\pm$0.55 & 69.34$\pm$0.47 & 70.84$\pm$0.58 & 70.60$\pm$0.62 & \textbf{73.51$\pm$0.52} & 72.27$\pm$0.59\\
\datasetnqopen(mistral) & 69.45$\pm$0.55 & 52.60$\pm$0.39 & 66.70$\pm$0.49 & 67.14$\pm$0.42 & 68.07$\pm$0.44 & \textbf{72.34$\pm$0.45} & 71.19$\pm$0.58\\
\bottomrule
\end{tabular}
}
\end{table*}
}

{
\setlength{\tabcolsep}{3pt}
\begin{table*}
\caption{
Like \cref{table:main:exp:arc:mlg}, but using accuracy from \LLaMATwoName.
}
\label{table:appendix:exp:arc:mlg:llama2}
\centering
   \resizebox{2\columnwidth}{!}{
\begin{tabular}{ccc|ccccc|cc}
\toprule
  & Random & Upper Bound & \baselineDegree(E) & \baselinePTrue & \baselineNLL & \baselineNLLNorm & \baselineSAR & \methodnamePrompt & \methodnameNext\\
\midrule
\datasettrivia(llama2) & 82.71$\pm$0.13 & 98.41$\pm$0.03 & 92.48$\pm$0.20 & 87.25$\pm$0.13 & 95.60$\pm$0.06 & 95.51$\pm$0.05 & 95.49$\pm$0.05 & \textbf{95.91$\pm$0.06} & 95.86$\pm$0.07\\
\datasettrivia(gemma) & 78.58$\pm$0.12 & 97.52$\pm$0.03 & 91.02$\pm$0.23 & 91.35$\pm$0.14 & 94.12$\pm$0.05 & 93.85$\pm$0.04 & 93.85$\pm$0.05 & \textbf{94.29$\pm$0.04} & 94.14$\pm$0.05\\
\datasettrivia(mistral) & 80.23$\pm$0.11 & 97.90$\pm$0.02 & 91.48$\pm$0.33 & 87.38$\pm$0.13 & 94.75$\pm$0.06 & 94.67$\pm$0.05 & 94.66$\pm$0.05 & \textbf{95.16$\pm$0.06} & \textbf{95.15$\pm$0.04}\\
\datasetcoqa(llama2) & 91.00$\pm$0.17 & 99.58$\pm$0.02 & 94.19$\pm$0.16 & 91.86$\pm$0.17 & 95.30$\pm$0.14 & 95.65$\pm$0.16 & 95.73$\pm$0.16 & \textbf{95.88$\pm$0.17} & 95.81$\pm$0.17\\
\datasetcoqa(gemma) & 92.14$\pm$0.15 & 99.68$\pm$0.01 & 95.10$\pm$0.26 & 93.39$\pm$0.15 & 96.09$\pm$0.11 & 96.12$\pm$0.14 & 96.18$\pm$0.13 & \textbf{96.35$\pm$0.12} & \textbf{96.39$\pm$0.14}\\
\datasetcoqa(mistral) & 91.61$\pm$0.16 & 99.64$\pm$0.01 & 94.81$\pm$0.34 & 92.28$\pm$0.18 & 95.47$\pm$0.11 & 95.63$\pm$0.13 & 95.71$\pm$0.12 & \textbf{95.86$\pm$0.18} & \textbf{95.86$\pm$0.15}\\
\datasetnqopen(llama2) & 56.74$\pm$0.60 & 88.89$\pm$0.34 & 68.96$\pm$1.00 & 57.57$\pm$0.60 & 68.96$\pm$0.70 & 69.01$\pm$0.73 & 69.72$\pm$0.71 & \textbf{71.10$\pm$0.64} & 70.80$\pm$0.78\\
\datasetnqopen(gemma) & 48.50$\pm$0.60 & 83.59$\pm$0.44 & 61.92$\pm$1.07 & 56.26$\pm$0.62 & 64.78$\pm$0.73 & 64.47$\pm$0.85 & 64.75$\pm$0.85 & \textbf{65.96$\pm$0.85} & 64.95$\pm$0.99\\
\datasetnqopen(mistral) & 53.65$\pm$0.59 & 87.05$\pm$0.37 & 65.99$\pm$0.99 & 56.07$\pm$0.86 & 67.63$\pm$0.69 & 67.29$\pm$0.57 & 68.31$\pm$0.54 & \textbf{69.88$\pm$0.64} & 69.46$\pm$0.54\\
\bottomrule
\end{tabular}
}
\end{table*}
}

\section{Calibration Quality}\label{appendix:calib}
We perform post-hoc calibration using the default \texttt{sklearn.calibration.CalibratedClassifierCV}.
\cref{fig:appendix:calib} shows the reliability diagrams  (similar to those in \citet{kull2019beyond}) after calibration for \methodnamePrompt.
The resulting probabilities are overall quite calibrated.

\def \FigCalibration{
\begin{figure*}[ht]
\centering
\includegraphics[width=0.8\textwidth]{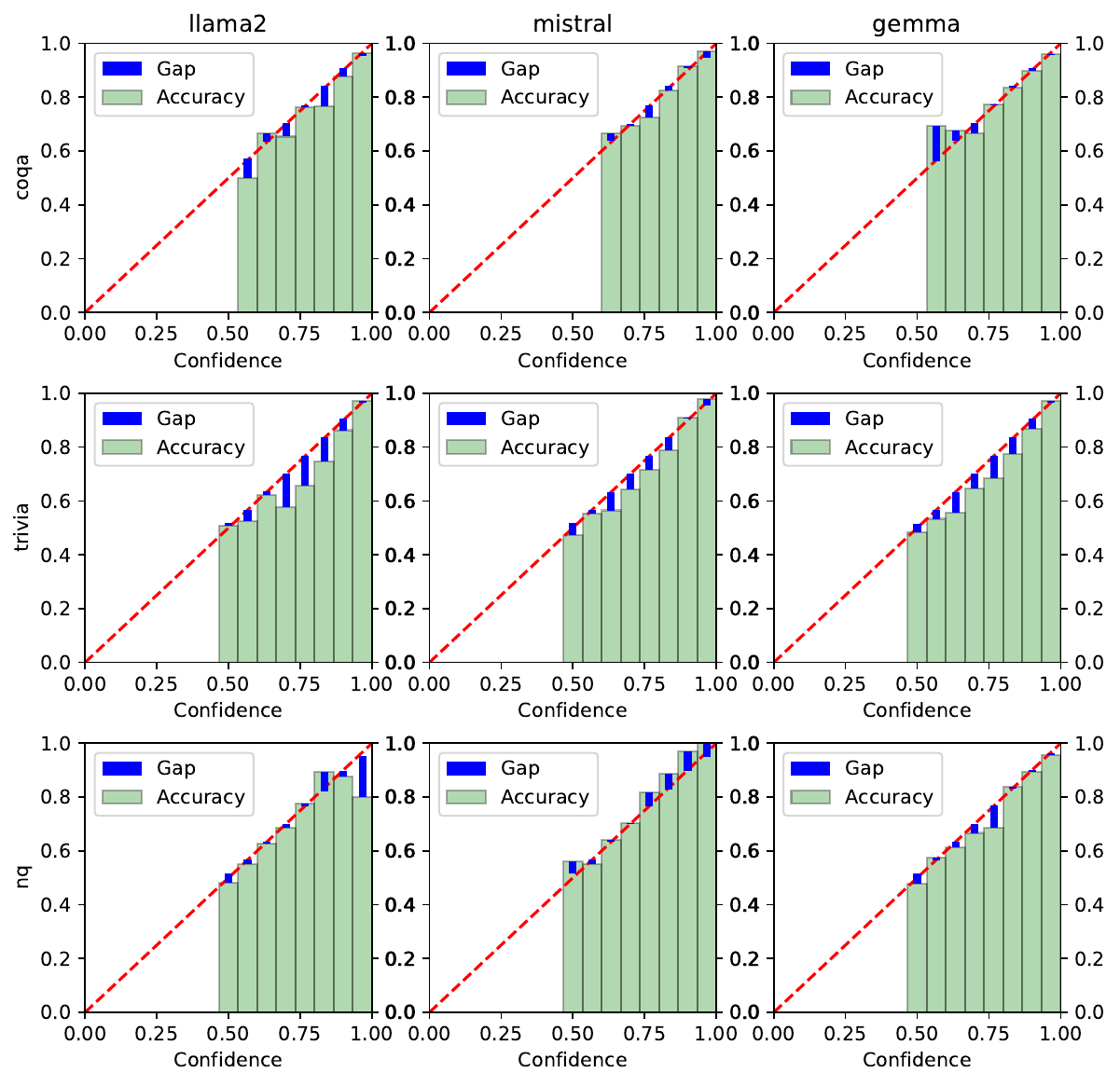}
\caption{
Reliability diagrams of \methodnamePrompt.
Bins with fewer than 10 samples are ignored due to noise.
}
\label{fig:appendix:calib}
\end{figure*}
}
\FigCalibration

\end{document}